\pdfoutput=1

\documentclass[11pt]{article}

\usepackage{naacl2021}

\usepackage{times}
\usepackage{latexsym}
\usepackage{amsmath}
\usepackage{graphicx}
\usepackage{subfigure}
\usepackage{caption}
\usepackage{xcolor}
\usepackage{booktabs}
\usepackage[T1]{fontenc}

\usepackage[utf8]{inputenc}

\usepackage{microtype}

\title{Evaluating the Impact of a Hierarchical Discourse Representation on Entity Coreference Resolution Performance}

\author{Sopan Khosla \quad James Fiacco \quad Carolyn Rose\\
	\vspace{-4 mm} \\
	Language Technologies Institute \\
	Carnegie Mellon University, USA \\
	{\tt \small \{sopank,jfiacco,cprose\}@cs.cmu.edu} \\
}

\begin{document}
\maketitle
\begin{abstract}
Recent work on entity coreference resolution (CR) follows current trends in Deep Learning applied to embeddings and relatively simple task-related features. SOTA models do not make use of hierarchical representations of discourse structure. In this work, we leverage  automatically constructed discourse parse trees within a neural approach and demonstrate a significant improvement on two benchmark entity coreference-resolution datasets. We explore how the impact varies depending upon the type of mention.
\end{abstract}

\section{Introduction}
Historically, theories of discourse coherence~\cite{chafe1976givenness,hobbs1979coherence,grosz1986attention,clark1991grounding} have offered elaborate expositions on how the patterns of anaphoric references in discourse are constrained by limitations in human capacity to manage attention and resolve ambiguity. \citet{hobbs1979coherence} acknowledges that these human limitations have meant that coreference resolution in natural text can be achieved with relatively high accuracy using a combination of recency and simple semantic constraints. State-of-the-art neural approaches for coreference resolution~\cite{lee2017end,joshi2019bert,joshi2020spanbert} have therefore not surprisingly shown strong performance relying on surface-level features and local-context (i.e., extracted from a small text window around the mention).  Traditional approaches, on the other hand, make an attempt to formally model the process of managing attention, for example, the stack in \citet{grosz1986attention}'s model. Their stack-based model suggests specific places where recency might fail while a more explicit model of discourse structure might make a correct prediction, for example, where an anaphor and a nearby potential (but incorrect) antecedent are in adjacent but separate discourse segments. Because of the potential existence of such cases, we hypothesize that formally incorporating a representation of discourse structure would have a small but non-random positive impact on the ability to correctly resolve anaphoric references.  This effect might vary depending upon the semantic informativeness of alternative types of anaphoric expressions, since they impose different constraints on where their antecedent can be located within a hierarchical discourse structure. There is also a danger that the level of accuracy with which the hierarchical structure of discourse can be obtained in practice might reduce the positive impact still further.  

The contribution of this paper is an empirical investigation of the impact of including a representation of the hierarchical structure of discourse within a neural entity coreference approach. To this end, we leverage a state-of-the-art RST discourse-parser to convert a flat document into a tree-like structure from which we can derive features that model the structural constraints. 
We embed this representation within an architecture that is enabled to learn to use this information deferentially depending upon the type of mention.  The results demonstrate that this level of nuance enables a small but significant improvement in coreference accuracy, even with automatically constructed RST trees.

\begin{figure*}[t]
    \centering
    \includegraphics[width=\textwidth]{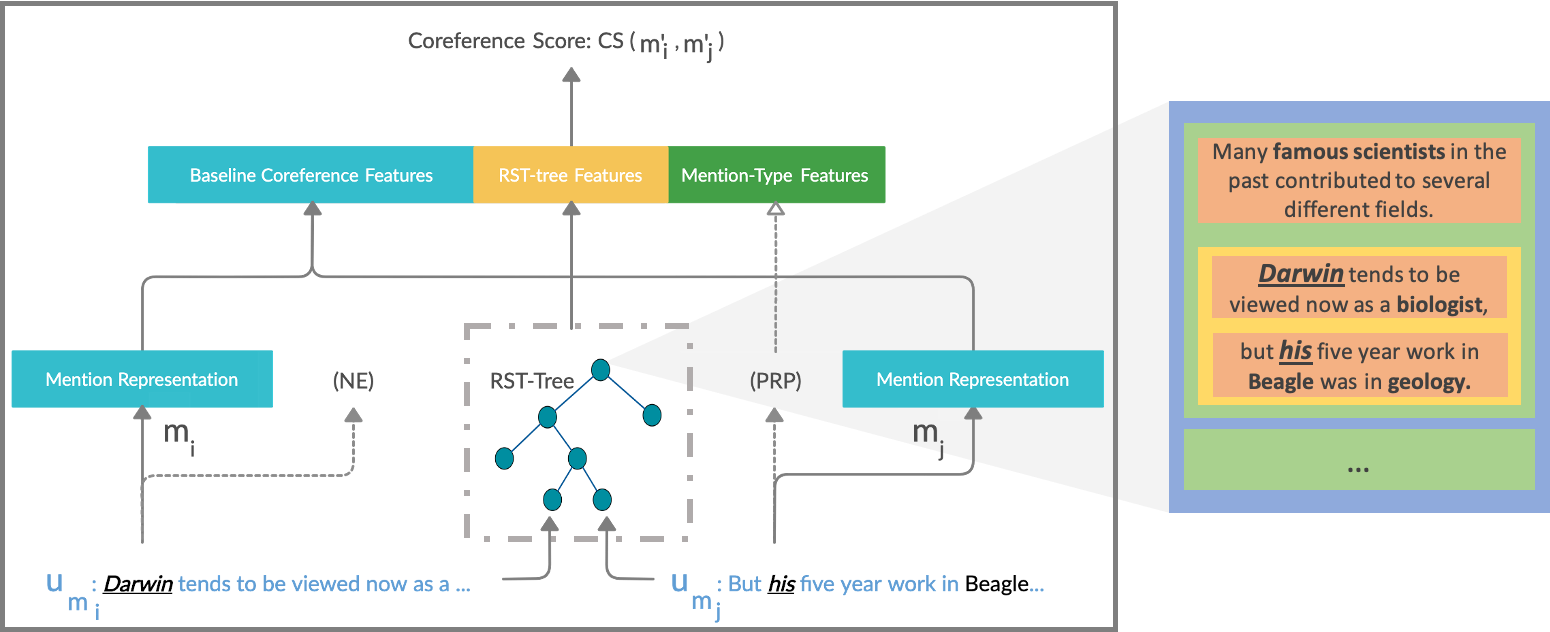}
    \captionsetup{font=small,labelfont=small}
    \captionof{figure}{Schematic diagram of 
    our discourse-informed neural architecture. 
    Discourse (yellow) and mention-type features (green) are concatenated with baseline features (blue) to obtain the mention-pair representation for scoring.
    }
    \label{fig:arch}
    \vspace{-0.3cm}
\end{figure*}
\section{Related Work}
Though recency is the strongest predictor for coreference resolution (CR), prior work in CR has benefited from the inclusion of semantic features such as type-information on top of the surface and syntax-level features. \citet{soon2001machine,bengtson2008understanding} used dictionaries like WordNet to extract the semantic class for a noun. 
More recently,~\citet{khosla2020using} showed that adding NER style type-information to \citet{lee2017end} substantially improves performance across multiple datasets. 

Discourse-level features have been successfully employed in multiple downstream NLP tasks like summarization~\cite{louis2010discourse}, sentiment analysis~\cite{somasundaran2009supervised}, and student writing evaluation~\cite{burstein2013holistic}. For coreference resolution, \citet{cristea1999discourse} showed that 
the potential of natural language systems to correctly determine co-referential links, can be increased by exploiting the hierarchical structure of texts. Their discourse model was informed by Vein Theory~\cite{fox_1987}, which 
identifies chains of elementary discourse units, over discourse structure trees that are built according to the RST \cite{mann1987rhetorical} requirements.
\citet{haghighi2010coreference} proposed an entity-centered model that leveraged discourse features like dependency-parse tree distance, sentence distance, and the syntactic positions (subject, object, and oblique) of
the mention and antecedent to perform coreference. 

In this work, we use \citet{parser}'s RST parser to convert documents into RST discourse-structure trees~\cite{mann1987rhetorical,taboada2006applications}. From these trees, we derive distance and coverage-based features to model the discourse-level structural constraints, which are passed as input to a neural-network based coreference resolver.
To our knowledge, ours is the first work that tries to explicitly incorporate discourse-level constraints for coreference resolution in a neural setting. 



\section{Model}
In this section, we explain how we introduce discourse-level features into a neural CR system.

\subsection{Baseline}
We leverage~\citet{lee2017end} as our baseline. We replace the word-embeddings with a BERT encoder. A preprocessing step for CR is to identify the mentions within the text that need to be resolved. Following~\citet{bamman2020annotated} and~\citet{khosla2020using}, we remove this possible source of error from our evaluation of entity coreference accuracy by using gold-standard mentions.




The baseline model's prediction of coreference for a pair of mentions, $\mathbf{S}\left(m_{i}, m_{j}\right)$, is computed as follows.  The representations of the two mentions $m_i$ and $m_j$ along with their element-wise product ($m_i \odot m_j$) and other features like distance between the mentions ($d_m$), and distance between the sentences that contain the mentions ($d_s$), are joined together and passed through a fully-connected layer (blue boxes in Figure~\ref{fig:arch}).

\begin{equation*}
\centering
\begin{split}
    mm_{ij} &= \left[m_{i} ; m_{j} ; m_{i} \odot m_{j} ; d_m ; d_s; ...\right] \\
    \mathbf{S}\left(m_{i}, m_{j}\right) &= \mathbf{FC}\left(mm_{ij}\right)
\end{split}
\end{equation*}

\subsection{Incorporating Discourse-level Features}
By incorporating a representation of the hierarchical discourse structure into the representation that is input to the neural model, we seek to add the capability for reasoning that is not possible in the baseline for each mention-pair ($mm_{ij}$).  None of the features included in the baseline distinguish between pairs that occur within the same or different discourse segments, for example.  The closest feature in the baseline that approximates document-level relationships is $d_s$, since it can be assumed that mentions are less likely to occur within the same segment the further apart they are in the discourse. But this is not universally true.

RST~\cite{mann1987rhetorical} offers a theoretical framework in which documents can be parsed into trees that capture the hierarchical discourse structure of the text. In this work, we incorporate structural features from such discourse trees, obtained automatically from \citet{parser}.
We concatenate three structural features, extracted from the discourse-tree of the document, with $mm_{ij}$ to model these constraints (as shown in Figure~\ref{fig:arch}). We use binarized RST-trees to represent the discourse hierarchy and relationships within each document. Discourse-units identified by the parser occur at the leaves ($l$) of the output tree.

Consider the document under consideration $doc$ and its RST-tree $t_{doc}$. For the current mention $m_j$ and candidate mention $m_i$, and the position of the smallest discourse-unit they belong to in the tree ($l_{u_{m_j}}$ and $l_{u_{m_i}}$ respectively):

\textbf{DistLCA ($\mathbf{d^j_{lca}}$)} encodes the distance between $l_{u_{m_j}}$ and $LCA(l_{u_{m_i}}, l_{u_{m_j}})$. This feature provides information about the amount of generality required to have the two mentions in the same discourse subtree. The smaller the \textit{DistLCA}, the closer the two mentions are assumed to be in the discourse.

\textbf{LeafCoverageLCA ($\mathbf{lc_{lca}}$)} encodes the number of sentences that are covered by the discourse subtree with $LCA(l_{u_{m_i}}, l_{u_{m_j}})$ as its root. This feature captures the coverage of the level of  discourse that encloses both mentions. The larger the \textit{LeafCoverageLCA}, the more the document area that needs to be covered to include both mentions.

\textbf{WordCoverageLCA ($\mathbf{wc_{lca}}$)} encodes the number of words that are covered by the discourse subtree with $LCA(l_{u_{m_i}}, l_{u_{m_j}})$ as its root. This feature is analogous to \textit{LeafCoverageLCA} but operates on word-level rather than the discourse-unit-level.


\subsection{Mention Types and Cognitive Load}
Across different types of anaphoric mentions, depending upon how much information about the antecedent is made apparent, there are differences with respect to the cognitive load imposed on the reader. Because this places differential constraints on the interpretation process, we hypothesize that enabling the model to learn different strategies depending upon the mention-type will be advantageous.  We divide mentions into three types ($type$) motivated by the above-mentioned intuition: (i) pronouns (low lexical information, high cognitive load on the reader), (ii) named-entities (already grounded mentions), and (iii) all other noun phrases. A mention is put in the second category if it contains at least one named-entity as predicted by an off-the-shelf NER system.\footnote{ https://demo.allennlp.org/named-entity-recognition} To identify pronouns, we compare the mention against a manually curated list of English pronouns. Ultimately, the discourse and mention-type features are concatenated with $mm_{ij}$ and passed through a fully-connected layer for scoring (Figure~\ref{fig:arch}).
\begin{equation*}
\centering
    \mathbf{S}(m_{i}, m_{j}) = \mathbf{FC}([mm_{ij};d^j_{lca}; lc_{lca}; wc_{lca}; type_j])
\end{equation*}

\section{Experimental Setup}
In this section, we describe the datasets and evaluation metrics we use in our experiments.
\subsection{Datasets}
We gauge the benefits of using RST-tree features on two state-of-the-art entity CR datasets discussed below.
Since, our off-the-shelf RST parser \cite{parser} is trained on news articles, the choice of datasets is motivated by the attempt at reducing the distribution shift between training and inference while ensuring that the parser was trained on different data than we are using for testing.  We use the English subset of \textbf{OntoNotes} \cite{onto2012}. The corpus contains multiple sub-genres ranging from news articles to telephone conversations.  We also evaluate our approach on a subset of the RST sub-genre of the ARRAU corpus \cite{arrau} (\textbf{A-RST(gt)}), which contains RST ground-truth parse-tree annotations in the RST Discourse-Treebank \cite{rst_treebank}.
Following~\citet{parser}, we keep 347 A-RST(gt) articles for training (out of which we set aside 22 articles for development), and 38 articles for testing. Although ARRAU also annotates bridging~\cite{clark1975bridge} and abstract anaphora~\cite{webber1991abstract}, in this work, we only focus on entity anaphora. 

\begin{table}[t]
    \resizebox{\linewidth}{!}{
        \centering
        \begin{tabular}{lcc}
        \toprule
            \textbf{Model} & \textbf{OntoNotes} & \textbf{A-RST(gt)} \\ 
        \midrule
            \textbf{\citet{lee2017end}} & 83.36 & 85.80\\
            \textbf{+ type} & \underline{83.70} & 85.95 \\
            \textbf{+ disc} & \underline{83.63} & \underline{86.19}\\
            \textbf{+ disc + type} & \underline{\textbf{83.89}} & \underline{\textbf{86.51}} \\
        \midrule
            \textbf{+ disc(gt)} & - & \underline{86.41} \\
            \textbf{+ disc(gt) + type} & - & \underline{\textbf{86.70}} \\
            + \textbf{disc(gt) + type} $\mathbf{- d_s}$& - & \underline{86.66} \\
        \bottomrule 
        \end{tabular}
    }
    \captionsetup{font=small,labelfont=small}
    \caption{Performance (Avg. F1) of discourse-informed model variants (gold-mentions) on OntoNotes and A-RST(gt). \underline{Underlined} numbers represent scores that are significantly different from the baseline ($p < 0.01$).\footnotemark}
    \label{tab:results}
    \vspace{-0.4cm}
\end{table}
\subsection{Evaluation Metrics}
Both OntoNotes and A-RST(gt) are input to the system in the CoNLL 2012 format. We evaluate the systems on the F1-score for MUC, B3, and CEAF metrics using the
CoNLL-2012 official scripts. However, we only show the average F1-score of the above-mentioned metrics in this paper for brevity.
We report the mean score of 5 independent runs with different seeds.\footnotetext{We leave the evaluation of the impact of including RST structural features in the
end-to-end CR setting as future work.}\footnote{Refer to Appendix A for hyperparameter values and Appendix B for detailed results.}

\section{Results}
\noindent\textbf{Ground-truth RST-Trees:}
To establish an upper-bound for the improvement through introduction of the discourse-tree features, we use features extracted from ground-truth trees. We evaluate the upper-bound performance on A-RST(gt) as it contains documents with annotations for coreference as well as RST-structures.
Our results show that incorporating ground-truth tree features along with the mention's type (\textbf{+ disc(gt) + type}) gives a boost of $0.90$ Avg. F1 ($p < 0.01$) over the baseline (Table~\ref{tab:results}), suggesting that discourse-level features are beneficial on A-RST(gt). Furthermore, we also find that
removing $d_s$ from this discourse-informed model does not cause a statistically-significant drop in performance. We believe that this happens because when discourse-structure features are included in the model, the signal from $d_s$ becomes redundant and sub-optimal.
\vspace{0.2cm}

\noindent\textbf{Predicted RST-Trees:}
In our second set of experiments we use discourse-trees extracted using \citet{parser}'s RST-parser. As shown in Table~\ref{tab:results}, adding predicted discourse-tree features improves over the baseline on both datasets, with A-RST(gt) corpus witnessing the highest absolute gain of $0.71$ Avg. F1 points. Please note that the results are statistically significant with $p < 0.01$.\footnote{We performed a one-tailed t-test to evaluate significance.}
The relative improvement on OntoNotes is smaller than A-RST(gt) ($0.53$ absolute Avg. F1 points). This could partially be explained by the fact that the RST-parser is trained on news articles, and therefore, might not generalize well on conversational sub-genres of OntoNotes like tc or bc.\vspace{0.2cm}

\noindent\textbf{Ablation Study:} To evaluate the contribution of each feature separately, we also perform an ablation study (Table \ref{tab:results}). On A-RST(gt), we find that the $type$ feature by itself does not provide a considerable boost over the baseline. 
Use of RST-tree based structural features, on the other hand, shows statistically significant improvements ($p < 0.01$), however, the jump is small (from $85.80$ to $86.19$). 
Our final model which includes both RT-tree features and $type$ gives the best results. \textbf{+ disc + type} performs much better than \textbf{+ disc} on both datasets (improvement of $0.32$ Avg. F1 points on A-RST(gt) and $0.26$ points on Onto) suggesting that the use of $type$ as a feature enhances the discriminative power of discourse-tree features. \vspace{0.2cm}
\begin{figure}[t]
    \centering
    \includegraphics[width=\linewidth]{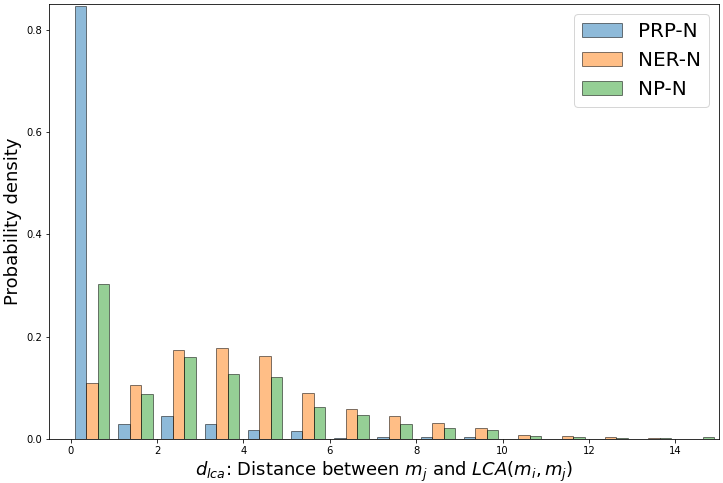}
    \captionsetup{font=small,labelfont=small}
    \caption{Distribution of $d_{lca}$ for the three different categories of anaphoric mention-pairs.}
    \label{fig:compare_all_norm}
    \vspace{-0.2cm}
\end{figure}

\noindent\textbf{Mention Type Analysis:} To study the influence of different mention-types on the discriminative power of discourse features, we analyze the distribution of $d_{lca}$ across different mention-pair categories in the A-RST(gt) training set.\vspace{0.2cm}

\emph{Setup.} To this end, we firstly extract relevant coreferent mention-pairs from the ground-truth clusters. To create a pair for each mention $m_j$, we choose the mention $m_i$ that belongs to the same cluster ($C$) as $m_j$, occurs before it in the document ($i < j$), and is the closest instance of $C$ to $m_j$.
Pairs created using this algorithm do not have other supporting mentions from the same cluster in between them.
We then extract three types of mention-pairs from these relevant pairs for our analysis: (i) $m_j$ is a pronoun and $m_i$ is not a pronoun (\textbf{PRP-N}); (ii) $m_j$ contains a named entity and $m_i$ is not a pronoun (\textbf{NE-N}); and (iii) $m_j$ is neither a pronoun nor contains a named entity, $m_i$ is not a pronoun, and $m_i, m_j$ have no lexical overlap (\textbf{NP-N}).\vspace{0.2cm}

\emph{Results.} Figure~\ref{fig:compare_all_norm} shows that there is indeed a dependence between $d_{lca}$ and mention-pair type. Most of the PRP-N pairs have a $d_{lca} < 5$ even though the the full RST-tree of a document can be as deep as $24$ levels. This corroborates our intuition that anaphors with higher ambiguity occur closer to their antecedents in the discourse. For NP-N, we find that $90\%$ of the pairs have $d_{lca} < 8$, whereas, $d_{lca}$ can go as large as $10$ for NE-N.
This trend explains, at least partially, the difference between the performance of discourse-informed models with and without the mention-type feature.
\section{Conclusion}
In this paper, we show that a representation of hierarchical discourse structure is beneficial for entity coreference resolution. Our proposed discourse-informed model observes small but statistically significant improvements over a state-of-the-art neural baseline on two coreference resolution datasets. Our analysis shows that the impact of the representation on performance is related to the cognitive load imposed by the type of anaphoric mention. 

While the model proposed in this work could serve as a useful baseline for the benefits of including discourse structure-based features in neural coreference resolution models, we realize that there is potential for achieving additional improvements by including more complex constraints (e.g. Right Frontier Constraint~\cite{asher2003logics}). We plan to study the affect of such features in future work.

\section*{Acknowledgements}

We thank the anonymous NAACL reviewers for
their insightful comments. We are also grateful to
the members of the TELEDIA group at LTI, CMU
for the invaluable feedback. This work was funded in part by NSF grants 1949110, 1822831, 1546393, and funding from Dow Chemical.

\bibliography{anthology,custom}
\bibliographystyle{acl_natbib}

\newpage

\appendix
\section*{Appendix}
\section{Hyperparameters}
\label{app:hp}

\begin{table}[!ht]
    \centering
    \resizebox{0.6\linewidth}{!}{
    \begin{tabular}{lr}
    \toprule
        \textbf{Hyperparameter} & \textbf{Value} \\
    \midrule
        BERT & base-cased \\
        BERT weights & freeze \\
        BiLSTM hidden dim & 200 \\
        $d_{lca}$ embedding-size & 20 \\
        $lc_{lca}$ embedding-size & 20 \\
        $wc_{lca}$ embedding-size & 20 \\
        $type$ embedding-size & 20 \\
        FC-layer 1 size & 150 \\
        FC-layer 2 size & 150 \\
        Dropout & 0.2 \\
    \bottomrule
    \end{tabular}
    }
    \captionsetup{font=small,labelfont=small}
    \caption{Hyperparameter values for our model. Reader is referred to \url{https://github.com/dbamman/lrec2020-coref} for the implementation of the baseline model.}
    \label{tab:hp}
\end{table}

\section{Detailed Results}
\label{app:detailed_res}

\begin{table}[h]
    \centering
    \resizebox{\linewidth}{!}{
        \begin{tabular}{l|ccc|ccc}
        \toprule
            \textbf{Model} & \multicolumn{3}{c|}{\textbf{OntoNotes}} & \multicolumn{3}{c}{\textbf{A-RST(gt)}} \\ 
        \midrule
            & \textbf{MUC} & $\mathbf{B^3}$ & \textbf{CEAF} & \textbf{MUC} & $\mathbf{B^3}$ & \textbf{CEAF} \\ 
        \midrule
            \textbf{\citet{lee2017end}} & 90.8 & 82.3 & 77.1 & 79.0 & 89.0 & 89.3 \\
            \textbf{+ type} & 91.1 & 82.7 & 77.3 & 79.3 & 89.0 & 89.4\\
            \textbf{+ disc} & 91.0 & 82.5 & 77.4 & 79.5 & 89.2 & 89.7 \\
            \textbf{+ disc + type} & 91.2 & 82.7 & 77.8 & 79.7 & 89.7 & 90.1 \\
        \midrule
            \textbf{- $d_s$} & - & - & - & 79.4 & 88.5 & 89.3\\
            \textbf{- $d_s$ + disc(gt) + type} & - & - & - & 80.2 & 89.6 & 90.2 \\
        \midrule
            \textbf{+ disc(gt)} & - & - & - & 79.8 & 89.5 & 89.9\\
            \textbf{+ disc(gt) + type} & - & - & - & 80.2 & 89.8 & 90.1 \\
        \bottomrule 
        \end{tabular}
    }
    \caption{Detailed Performance (F1 score) of discourse-informed model variants (gold-mentions) on OntoNotes and A-RST(gt).}
    \label{tab:detailed_results}
\end{table}

\section{Results on Validation Set}
\label{app:val_res}

\begin{table}[h]
    \centering
    \resizebox{0.8\linewidth}{!}{
        \centering
        \begin{tabular}{lcc}
        \toprule
            \textbf{Model} & \textbf{OntoNotes} & \textbf{A-RST(gt)} \\ 
        \midrule
            \textbf{\citet{lee2017end}} & 83.42 & 85.98\\
            \textbf{+ type} & \underline{84.01} & 86.15 \\
            \textbf{+ disc} & \underline{83.91} & \underline{86.40}\\
            \textbf{+ disc + type} & \underline{\textbf{84.38}} & \underline{\textbf{86.74}} \\
        \midrule
            \textbf{+ disc(gt)} & - & \underline{86.68} \\
            \textbf{+ disc(gt) + type} & - & \underline{\textbf{87.02}} \\
        \bottomrule 
        \end{tabular}
    }
    \caption{Performance (Avg. F1) of discourse-informed model variants (gold-mentions) on OntoNotes and A-RST(gt) validation set.}
    \label{tab:val_results}
\end{table}

\section{Computational Infrastructure}
\label{app:infra}

All our experiments are performed on a single Nvidia GeForce GTX 1080 Ti GPU. Training took 20-22 hours on OntoNotes, and 5-7 hours on A-RST(gt). We trained the models for 100 epochs with an early-stopping criteria on the Avg. F1 performance on the validation set.

\end{document}